\documentclass[twocolumn]{autart}    
\pdfoutput=1 
\usepackage{graphicx}      
\usepackage{mathptmx}
\usepackage{amsmath,amsfonts,amssymb}
\usepackage{courier}    
\usepackage[numbers]{natbib}
       
\usepackage{color}
\usepackage{subcaption}
\usepackage{parskip}
\usepackage{multirow}
\usepackage{caption}
\usepackage{svg}
\usepackage{url}
\usepackage{hyperref}
\usepackage[T1]{fontenc}

\begin{document}

\begin{frontmatter}

\title{Port-Hamiltonian Neural Networks with Output Error Noise Models\thanksref{footnoteinfo}} 

\thanks[footnoteinfo]{This work is part of the DAMOCLES research project which received funding from the Eindhoven Artificial Intelligence Systems Institute, as part of the EMDAIR funding programme. Furthermore, this work is funded by the European Union within the framework of the National Laboratory for Autonomous Systems (RRF-2.3.1-21.2022-00002).}

\author[First,Second]{Sarvin Moradi} \ead{s.moradi@tue.nl},
\author[First] {Gerben I. Beintema} \ead{g.i.beintema@tue.nl},
\author[Second,Third]{Nick Jaensson} \ead{n.o.jaensson@tue.nl}, 
\author[First,Fourth]{Roland T\'{o}th} \ead{r.toth@tue.nl},
\author[First,Second]{Maarten Schoukens} \ead{m.schoukens@tue.nl}

\address[First]{Control System group, Eindhoven University of Technology, Eindhoven, the Netherlands}
\address[Second]{Eindhoven Artificial Intelligence Systems Institute, Eindhoven, the Netherlands}
\address[Third]{Processing and Performance of Materials Group, Eindhoven University of Technology, Eindhoven, the Netherlands}
\address[Fourth]{Systems and Control Laboratory, HUN-REN Institute for Computer Science and Control, Budapest, Hungary}


\begin{abstract}                          
Hamiltonian neural networks (HNNs) represent a promising class of physics-informed deep learning methods that utilize Hamiltonian theory as foundational knowledge within neural networks. However, their direct application to engineering systems is often challenged by practical issues, including the presence of external inputs, dissipation, and noisy measurements. This paper introduces a novel framework that enhances the capabilities of HNNs to address these real-life factors. We integrate port-Hamiltonian theory into the neural network structure, allowing for the inclusion of external inputs and dissipation, while mitigating the impact of measurement noise through an output-error (OE) model structure. The resulting output error port-Hamiltonian neural networks (OE-pHNNs) can be adapted to tackle modeling complex engineering systems with noisy measurements. Furthermore, we propose the identification of OE-pHNNs based on the subspace encoder approach (SUBNET), which efficiently approximates the complete simulation loss using subsections of the data and uses an encoder function to predict initial states. By integrating SUBNET with OE-pHNNs, we achieve consistent models of complex engineering systems under noisy measurements. In addition, we perform a consistency analysis to ensure the reliability of the proposed data-driven model learning method. We demonstrate the effectiveness of our approach on system identification benchmarks, showing its potential as a powerful tool for modeling dynamic systems in real-world applications. 
\end{abstract}
\end{frontmatter}

\section{Introduction}

System identification is the process of developing mathematical models for dynamic systems given input-output data collected from real-world observations \cite{ljung2010perspectives}. This field utilizes various methods, including linear, nonlinear, and hybrid models, to address different levels of system complexity. Nonlinear system identification becomes essential when linear methods fall short, particularly due to the nonlinear and time-varying behaviors of real-world systems. As a result, there is a growing need for advancements in nonlinear system identification techniques \cite{nonlinear_SI}. 

Various techniques have been proposed for black-box nonlinear system identification over the last decade, including linear parameter-varying methods \cite{toth2010modeling}, Volterra series \cite{cheng2017volterra}, nonlinear ARMAX (NARMAX) models \cite{billings2013nonlinear}, and nonlinear state-space approaches \cite{schon2011system, beintema2023continuous} to mention a few. In this paper, we focus on nonlinear state-space identification approaches that incorporate prior physical knowledge. 
We aim to introduce physics into the modeling process to enable flexible parameterization and ensure physical relevance. 
In this context, we focus on port-Hamiltonian systems, which are particularly well-suited for modeling a wide range of systems across various fields.

A key advantage of port-Hamiltonian systems is their inherent stability, as their passivity ensures a solid foundation for maintaining stability in the behaviour of system  \cite{port_Hamilton}. This stability is particularly valuable when addressing complex input-output mappings, which can be a significant challenge in system identification \cite{stabilty_2024}. Additionally, the inherent compositionality of port-Hamiltonian systems makes them ideal for modular modeling,  allowing for the integration of multiple subsystems and effectively modeling more complex, large-scale systems.  Despite these advantages, there remains a gap in consistent data-driven approaches that combine these benefits with efficient system identification techniques.

In recent years, Hamiltonian neural networks (HNNs) have emerged as a promising approach which merge theoretical physics with data-driven approaches. HNNs are particularly helpful in modeling energy-conservative systems. Choudhary et al. successfully applied HNNs to simulate systems with chaotic behavior and conserved energy \cite{chaos2020}. Additionally, adaptable HNNs, introduced by Han et al., predict the behavior of nonlinear physical systems at varied parameter values \cite{adaptable}. 

Although these applications have been successful, HNNs face limitations in modeling non-energy conservative systems with inputs and/or dissipation. 
Sosanya and Greydanus proposed Dissipative Hamiltonian Neural Networks (D-HNNs) with introducing Rayleigh dissipation function to account for dissipation \cite{sosanya2022dissipative}. Zhong et al. generalized HNNs for Hamiltonian systems with external inputs \cite{Zhong2020Symplectic}; 
To address modeling systems with both input and dissipation, dissipative symODEN was introduced by Zhong et al. \cite{zhong2020dissipative}. This approach relied on a simplified version of port-Hamiltonian theory for mechanical systems with specific assumptions on the dissipation and input forces. 
Desai et al. developed port Hamiltonian neural networks (p-HNNs) to learn explicit time-dependent dynamical systems \cite{desai2021port}, adapting the formulation in \cite{zhong2020dissipative} by modeling input forces as neural networks with time as input and force as output, while treating dissipation as a learnable parameter. Most recently, Neary and Topcu used the compositional properties of the port-Hamiltonian theory for modeling complex systems \cite{port_HNN}. Their work significantly generalized the port-Hamiltonian framework beyond \cite{zhong2020dissipative} and \cite{desai2021port}. They parametrized the dissipation, input matrices, and Hamiltonian functions as neural networks while incorporating their mathematical properties. They trained the neural network with prediction loss function, given measurements of all states, and inputs. Although, in real world-engineering systems, measurement of the entire state vector is often not available and when available it is often noisy.

Despite of all these remarkable advancements, challenges still remain in the application of port-Hamiltonian neural networks. Ensuring consistency is crucial to guarantee that the identified models accurately represent the true system dynamics. Additionally, there is an emerging need for computationally efficient methods capable of handling large datasets and complex systems with partial state measurments. Furthermore, effectively addressing noisy measurements is critical for successfully using this approach in real-world applications.

To address these issues, this work introduces output-error port-Hamiltonian neural networks (OE-pHNNs). By introducing port-Hamiltonian theory into the neural network structure, our approach accommodates external inputs, dissipation, and handels measurement noise through an output-error (OE) model. To estimate OE-pHNNs, an identification approach using a subspace encoder structure, which is based on the extension of SUBNET \cite{beintema2023continuous} is introduced.  This approach efficiently approximates simulation loss using short simulations and an encoder function to predict initial states. A preliminary version of this work focusing on OE-HNNs with inputs appeared in \cite{moradi2023physics}.

Our main contributions to this work can be summarized as follows:
\begin{itemize}

 \item  \textit{(Identification method)} We develop the OE-pHNN approach and introduce a customized, efficient identification method for it using SUBNET.

 \item (\textit{Handling noisy measurements}) The identification approach ensures unbiased parameter estimates by maintaining statistical efficiency under the given noise settings.

 \item \textit{(Consistency)} We prove the consistency of the proposed method, ensuring its reliability in capturing complex system dynamics. 

\end{itemize}

\begin{figure*}
    \centering
    \includegraphics[width=0.75\linewidth]{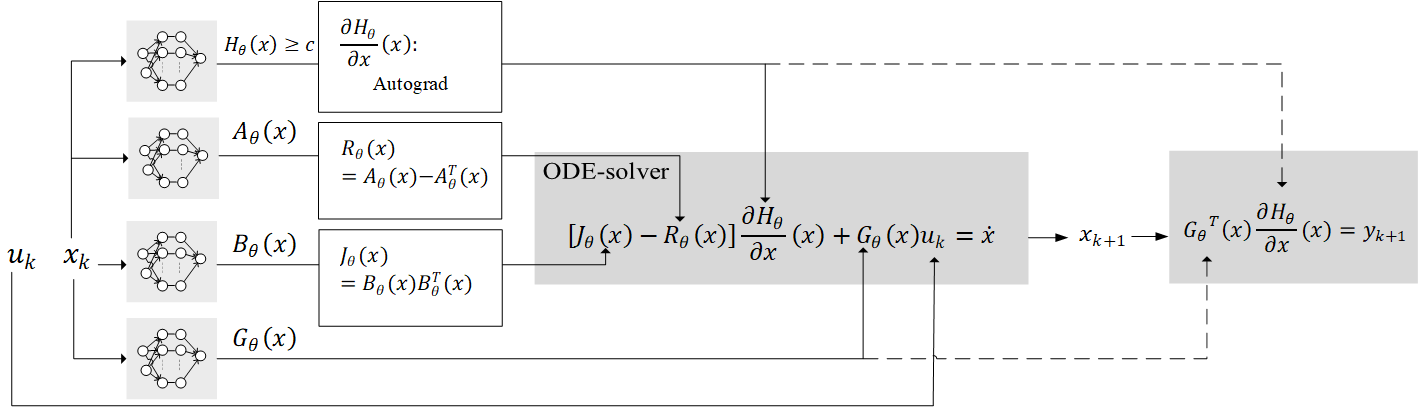}
    \caption{Output error port-Hamiltonian neural network (OE-pHNN).}
    \label{fig:port-HNN}
\end{figure*}

\section{Problem setting and preliminaries}
\subsection{Port-Hamiltonian systems} \label{subsec:Port-Hamiltonian systems}
Port-Hamiltonian theory enables the systematic representation and analysis of complex systems by incorporating energy conservation principles and structural considerations.
Central to this theory is the representation of the overall energy of the system using a Hamiltonian function ($H$), and the power exchange between the system and its ports which is described by a set of equations known as the input-state-output equations \cite{port_Hamilton}:
\vspace{-0.25cm}\begin{subequations}\label{eq:ss-portHNN} 
\begin{align} 
    \dot{{x}}(t) &= \overbrace{[{J(x(t))-R(x(t))}] \frac{\partial H}{\partial {x}}(x(t)) + G(x(t))u(t)}^{f({x}(t), {u}(t))},\\
    y_0(t) &= \underbrace{G^\intercal(x(t))\frac{\partial H}{\partial {x}}(x(t))}_{h({x}(t))} ,
\end{align}
\end{subequations} 
where $x(t)\in\mathbb{R}^{n}$ is the state, $u(t)\in\mathbb{R}^{m}$ is the input, and $y_0(t)\in\mathbb{R}^{m}$ is the corresponding output signal. The function $f$ governs the evolution of state ${x}$, capturing the dynamics of the system, and $h$ is the output function. $H(x) \in \mathbb{R}$ is a scalar function that denotes the total energy of the system. $J(x)\in\mathbb{R}^{n\times n}$ is a skew-symmetric function (i.e., $J: \mathbb{R}^n \rightarrow \mathbb{R}^{n\times n}, J(x)=-J^\top(x), \forall x\in\mathbb{R}^n $) which ensures the energy conservation in storage elements. $R(x)\in\mathbb{R}^{n\times n}$ describes energy dissipation of the system as a symmetric positive semi-definite state-dependent matrix  (i.e., $R(x)=R^\intercal(x), \forall x: x^\intercal R(x)x\geqslant 0$), and $G(x)\in\mathbb{R}^{n\times m}$ is the input matrix which shows how the input energy enters into the system. 
Port-Hamiltonian systems are inherently (cyclo-)passive. This property is implied by the conditions of boundedness of the Hamiltonian function $H(x)$ from below (i.e., $H(x)\geq c$), skew-symmetricity of the interconnection matrix $J(x)$, and positive semi-definiteness of the dissipation matrix $R(x)$. Under these prerequisites, the differential dissipation inequality \eqref{eq:passivity} is fulfilled:
\vspace{-0.25cm}
\begin{equation}\label{eq:passivity} 
    \frac{dH}{dt} = \frac{\partial H}{\partial {x}}^T J(x) \frac{\partial H}{\partial {x}} - \frac{\partial H}{\partial {x}}^T R(x) \frac{\partial H}{\partial {x}} + \frac{\partial H}{\partial {x}}^T G(x) u \leq y^T u.
\end{equation}
Here $H$ represents the storage function, while $u$, and $y$ denote the supply port variables.

\subsection{Considered systems} \label{subsec: considered system}
By rewriting \eqref{eq:ss-portHNN} and including noise in output measurements, we have
\begin{subequations}\label{eq:system_class} 
    \begin{align} \label{subeq:state_time derivative system}
        \dot{x}(t) &=f(x(t),u(t)) ,\\  \label{subeq:output function}
        y_0 (t) &= h(x(t)), \\
        y_k &= y_0 (kT_\mathrm{s}) + {v}_k; \label{subeq:sampled measurements}
    \end{align}
\end{subequations}
where, the function $f$ governs the evolution of state $x$, capturing
the dynamics of the system and $h$ is the output function as introduced in \eqref{eq:ss-portHNN}. The output measurements ${y}_k$ are sampled with sampling rate $T_\mathrm{s}$. 
These measurements are contaminated by ${v}_k$, a zero-mean white noise with finite variance, independent of $u(t)$. With $u_k=u(kT_\mathrm{s})$, we denote the sampled input-output pairs as 
 \begin{equation}\nonumber 
 {\mathcal{D}_N=\{({u}_k,{y}_k)\}_{k=0}^{N-1}}.
 \end{equation}
 
\subsection{Discrete-time system}\label{subsec:Discretized-time system}
Here we are interested in computing the sampled output ($y_k$) as introduced in \eqref{subeq:sampled measurements}, given the discrete time input $u_k$, which is transformed to continuous time by using the ZOH transformation. By recalling \eqref{subeq:state_time derivative system}, one can compute the state given a known initial state, $x_0$, as
\begin{equation} \label{eq: integral_state}
    x (t) = x_0 + \int_{0}^{t} f(x (t),u(t))\,dt.
\end{equation}
By integrating over one sampling period $T_\mathrm{s}$ and with ZOH assumption on inputs \cite{pintelon2012system}, we can directly compute the states for the subsequent sampling instant via
\begin{equation}\label{eq:integral_system}
    x((k+1)T_\mathrm{s}) = x (kT_\mathrm{s}) +\underbrace{\int_{0}^{T_\mathrm{s}} f(x (kT_\mathrm{s}+\tau),u(kT_\mathrm{s}))\,d\tau}_{I(f,T_\mathrm{s},x (kT_\mathrm{s}),u(kT_\mathrm{s}))},
\end{equation}
We approximate $I(f,T_\mathrm{s},x_k,u_k)$ in \eqref{eq:integral_system} with an $m^{th}$-order single-step numerical integrator, i.e. for $m\geq 1$:
\begin{equation} \label{eq:I(f)}
    I_m(f,T_\mathrm{s},x_k,u_k)  = \sum_{i=1}^{m} w_{i,m} f(x_{i,m}, u_k),
\end{equation}
where $w_{i,m}$ are integration weights, and $x_{i,m}$ are the integration nodes within $[0,T_\mathrm{s}]$; $m$ represents the number of integration nodes used in the approximation and $x_k=x(kT_\mathrm{s})$. The interval $[0,T_\mathrm{s}]$ is divided into $m$ subintervals, with the weights and nodes determined by the numerical integration method. The nodes $x_{i,m}$ represent the states evaluated at times $t_{i,m} = (k +c_{i,m}) T_\mathrm{s}$, where $c_{i,m} \in [0,1]$ are normalized time nodes determined by the chosen numerical integration formula. It is assumed that the chosen numerical integrator is convergent \cite{noren2022numerical}, and as the order $ m $ increases, the approximation of the integral in \eqref{eq:I(f)} becomes more accurate. 

\begin{thm} \label{thm:Convergence of numerical integration} (\textit{Convergence of numerical integration}). Let the state vector be defined as  
$x_k = [x_{1,k} \; x_{2,k} \; \dots  \; x_{n,k}]^\top,$ where $ x_{j,k}$ represents the $ j $-th element of $ x_k $. For the function $ f: \mathbb{X} \times \mathbb{U} \to \mathbb{R}^n $, $f_j(x,u)$ denotes the $j$-th component of $f (x,u)$. Similarly, the $ j $-th component of the  numerical integral in \eqref{eq:I(f)} is  
\begin{equation} 
 I_m(f, T_\mathrm{s}, x_k, u_k)_j = \sum_{i=1}^{m} w_{i,m} f_j(x_{i,m}, u_k),
\end{equation}
where $ f_j(x,u) $ denotes the $ j$-th component of $f(x,u)$ \cite{butcher}. 

 For a smooth vector-valued function $ f: \mathbb{X} \times \mathbb{U} \to \mathbb{R}^n $, the numerical integration satisfies:  
\begin{equation}
\lim_{m \to \infty} \sup_{(x_k, u_k) \in \mathbb{X} \times \mathbb{U}} \max_{1 \leq j \leq n} \frac{|I(f, T_\mathrm{s}, x_k, u_k)_j - I_m(f, T_\mathrm{s}, x_k, u_k)_j|}{|I(f, T_\mathrm{s}, x_k, u_k)_j|} = 0.
\end{equation}  

This result holds for $ T_\mathrm{s} < 1 $, assuming $ f(x, u) $ is sufficiently smooth, $ \mathbb{X} \times \mathbb{U} $ is compact, and the numerical integration method is convergent in the sense of \cite{atkinson1991introduction}.
\end{thm}

\textbf{Proof.}  
Denote the integration error for the $ j $-th component of $I_m(f,T_\mathrm{s},x_k,u_k)$ as:  
\begin{equation}
E_m(f, T_\mathrm{s}, x_k, u_k)_j = I(f, T_\mathrm{s}, x_k, u_k)_j - I_m(f, T_\mathrm{s}, x_k, u_k)_j.
\end{equation}  
For convergent numerical integrators of order $m$, such as the Runge-Kutta methods  
\begin{equation}
E_m(f, T_\mathrm{s}, x_k, u_k)_j = O(T_\mathrm{s}^{m+1}),
\end{equation}  
uniformely valid for $ m \geq 1 $ \cite{atkinson1991introduction}.  

For small sampling periods $ T_\mathrm{s} $ and nonzero $ f $, the integral $I(f,T_\mathrm{s},x_k,u_k)$ satisfies:  
\begin{equation}
|I(f, T_\mathrm{s}, x_k, u_k)_j| = O(T_\mathrm{s}),
\end{equation}  
for any $(x_k, u_k) \in \mathbb{X} \times \mathbb{U} $, since $f(x, u)$ is smooth and bounded over the compact domain $ \mathbb{X} \times \mathbb{U} $. Thus, the normalized error becomes:  
\begin{equation} \label{eq:component_error}
\frac{|E_m(f, T_\mathrm{s}, x_k, u_k)_j|}{|I(f, T_\mathrm{s}, x_k, u_k)_j|} = O(T_\mathrm{s}^m),
\end{equation}  
which converges uniformly to 0 over a compact $\mathbb{X} \times \mathbb{U}$ when $m \to \infty$. Note that when $f(x_k,u_k) = 0$, the numerator in the normalized error \eqref{eq:component_error} vanishes faster with $T_\mathrm{s}\rightarrow 0$ than the denominator. $\quad \blacksquare$

By introducing \eqref{eq:I(f)} into \eqref{eq:integral_system}, we have
\begin{equation}\label{eq:ODE_system}
    x((k+1)T_\mathrm{s}) \approx x (kT_\mathrm{s}) + I_m(f,T_\mathrm{s},x (kT_\mathrm{s}),u (kT_\mathrm{s})).
\end{equation}
We can rewrite \eqref{eq:ODE_system} as
\begin{equation} \label{eq:one-step ahead}
    x_{k+1} = {f_{d,m}} (x_k, u_k); 
\end{equation}
${f_{d,m}}$ takes the current sampled state, $x_k$, and the corresponding input, $u_k$, as inputs and allows us to calculate the next state, $x_{k+1}$. Given a known initial state and sampled inputs, the function ${f_{d,m}}$ facilitates the computation of the entire sequence of sampled states.

Having computed the sampled states, we can apply \eqref{subeq:output function}, and \eqref{subeq:sampled measurements} to compute the sampled outputs. For $n \geq 1$, the sampled output $y_k$ is given by
\begin{subequations}\label{eq:recurrent}
\begin{align}
     & y_k = h(x_k) + v_k,\label{eq:1st step_hf}\\
     & y_{k+1} = h\circ {f_{d,m}} (x_k, u_k^k) + v_{k+1},\label{eq:2nd step_hf}\\
     & \begin{aligned}
        & \quad\vdots \\
        & y_{k+n} = h\circ_n {f_{d,m}} (x_k, u_k^{k+n-1}) + v_{k+n},\label{eq:nth step_hf}
    \end{aligned}
\end{align}
\end{subequations}
where, the symbol $\circ$ represents function concatenation, i.e. it gives $ h(f_{d,m}(x_k,u_k))$ for $\circ_1$. Additionally, $\circ_n$ denotes the recursive repetition of $\circ$, $n$ times. For instance, $h \circ_2 {f_{d,m}} = h \circ {f_{d,m}} \circ {f_{d,m}}$. Here, $u_k^{k+n-1} = \begin{bmatrix} u_k^\intercal & ... & u_{k+n-1}^\intercal \end{bmatrix} ^\intercal$. With $y_k^{k+n}$ and $ v_k^{k+n}$ defined in the same way, one can rewrite \eqref{eq:recurrent} in a compact equation as
\begin{equation} \label{eq:compact}
    y_k^{k+n} = \Gamma_{n,m} (x_k, u_k^{k+n-1}) + v_k^{k+n}.
\end{equation}
As the sequence of noise samples, $v_k^{k+n}$, is unavailable, direct computation of the outputs using \eqref{eq:compact} is not feasible. If we compute the conditional expectation of \eqref{eq:compact} assuming $v_k$ as white noise with zero mean and independent of $u_k$, we will have
\begin{equation} \label{eq:1_step ahead predictor}
\begin{aligned}
        \bar{y}_{k}^{k+n} &= \mathbb{E}[y_{k}^{k+n} \mid x_k, u_k^{k+n-1}] \\
        &=\Gamma_{n,m}(x_k, u_k^{k+n-1});
\end{aligned}
\end{equation}
For a specific sample realization, we can write $\bar{y}_{n} = \gamma_{n,m} (x_1, u_1^{n-1}) $ with $\gamma_{n,m} = (h\circ_n {f_{d,m}} )$.

Recalling Equation \eqref{eq:compact}, for $n\geq 1$ and partially invariable $\Gamma_{n,m}$ w.r.t $x_k$, we can define the observabilty map as
\begin{equation}\label{eq:observabilty map}
    x_k = \phi_n(u_k^{k+n-1},y_k^{k+n}- v_k^{k+n}).
\end{equation}
Now we can build the reconstructability map, $\psi_n$ which allow us to extract $x_k$ from the past $n$ measured input and output data via
\begin{subequations}\label{eq:reconstructabilty map}
\begin{align}
        x_k &= (\circ_n {f_{d,m}})(x_{k-n}, u^{k-1}_{k-n})  \\ \label{subeq:phi}
            &= (\circ_n {f_{d,m}})(\phi_n(u^{k-1}_{k-n}, y^{k}_{k-n}- v_{k-n}^{k}), u^{k-1}_{k-n}) \\ 
            &= \psi_n(u^{k-1}_{k-n}, y^{k}_{k-n}-v_{k-n}^{k}).\label{subeq:psi}
\end{align}
\end{subequations}  
Also, given that $v_k$ is a zero-mean white noise with finite variance, we have
\begin{equation} \label{eq:estimated reconstructability map}
    \bar{x}_k = \mathbb{E} [x_k\mid u_{k-n}^{k-1}, y_{k-n}^k] = \bar\psi_n (u_{k-n}^{k-1}, y_{k-n}^k).
\end{equation}

\section{Identification approach}
\subsection{Model class}\label{subsec:Model class}
For the identification of the continuous system \eqref{eq:system_class} via collected data $\mathcal{D}_N$, we need to estimate the $f$ and $h$ functions. 
In order to model the system \eqref{eq:system_class}, one can employ parameterized functions $f_\theta$ and $h_\theta$ as introduced in \eqref{subeq:f_theta} and \eqref{subeq:h_theta}, respectively. Note that here an OE model structure is selected for modeling. \vspace{-0.25cm}
\begin{subequations}\label{eq: port_H_parametrized} 
\begin{align} \label{subeq:f_theta}
    \hat{\dot{x}} (t) &= f_{\theta}(\hat{{x}}(t), {u}(t)) \\   \nonumber 
    &= [(J_{\theta}(\hat{x}(t)) - R_{\theta}(\hat{x}(t))]\frac{\partial H_{\theta}}{\partial \hat{{x}}}(\hat{x}(t))+ G_{\theta}(\hat{x}(t)) u(t),\\ \label{subeq:h_theta}
    \hat{{y}}(t) &= h_{\theta}(\hat{{x}}(t)) = G_{\theta}^\intercal(\hat{x}(t))\frac{\partial H_{\theta}}{\partial \hat{{x}}}(\hat{x}(t)),\\ 
    \hat{y}_k &= \hat{y}(kT_\mathrm{s}),
\end{align}
\end{subequations}
where, $u(t)$ is under ZOH actuation, i.e. $u(t) = u(kT_\mathrm{s}), \forall t\in [kT_\mathrm{s}, (k+1)T_\mathrm{s} )$. In \eqref{eq: port_H_parametrized}, the terms with subscript $\theta$ are considered to be functions
parameterized with a parameter vector $\theta \in \Theta \subseteq \mathbb{R}^{n_\theta}$. These
functions are formulated as multi-layer feedforward neural networks (MLPs) in this paper.

The properties of the matrices \(J_\theta(x)\), \(R_\theta(x)\), and \(H_\theta(x)\) introduced in Subsection \ref{subsec:Port-Hamiltonian systems}, must be carefully taken into account when parameterizing these matrices. 
We define \(R_\theta\) as the result of multiplying matrix function \(A_\theta\) by its transpose, \(R_\theta = A_\theta {A_\theta}^T\). This formulation represents a specific matrix structure allowing for a symmetric positive semi-definite matrix \(R_\theta\). We also express \(J_\theta\) as the difference between a matrix function \(B_\theta\) and its transpose, denoted as \(J_\theta = B_\theta - {B_\theta}^T\). This representation guarantees that \(J_\theta\) possesses the properties of a skew-symmetric matrix. To enforce cyclo-passivity in the results, $H_\theta$ is constrained by a user-defined constant as the lower bound. We use the Exponential Linear Unit (ELU) activation function in the final layer of the neural network to maintain this lower bound for $H_\theta$ . Fig. \ref{fig:port-HNN} illustrates the structure of these matrices and their contribution to the modeling process. 

\begin{figure*}
    \centering
    \includegraphics[width=0.75\linewidth]{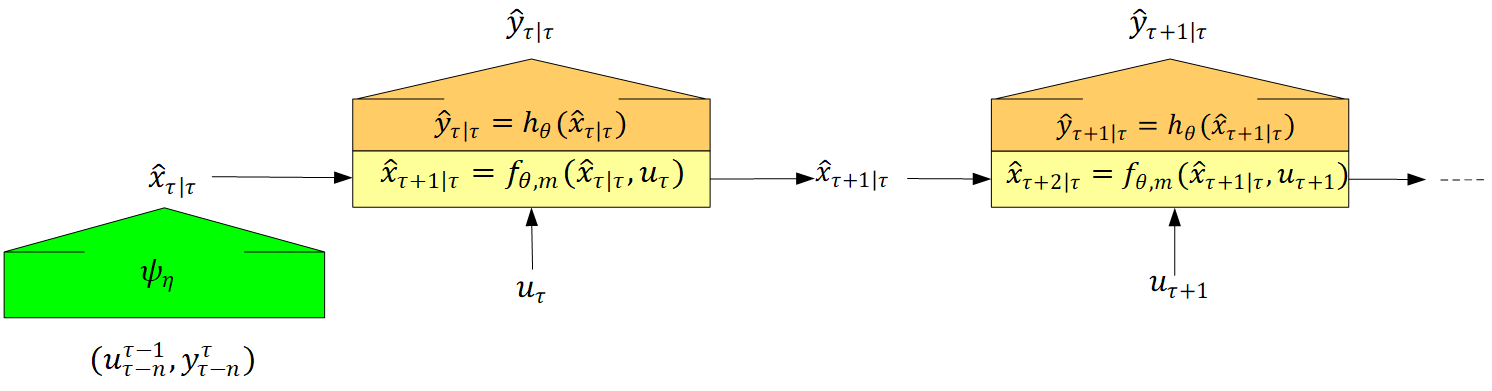}
    \caption{OE-pHNN and SUBNET: the subspace encoder $\psi_\eta$ estimates the initial state using past inputs and outputs. Then, the state is iteratively propagated through the framework until it reaches the truncation length $T$.}
    \label{fig:port-HNN and SUBNET}
\end{figure*}


\subsection{Discrete-time form}\label{subsec:Discretized-time model}
We can argue that similar to the discrete-time system description in Section \ref{subsec:Discretized-time system} and Equation \eqref{eq:ODE_system}, a discrete-time version of \eqref{subeq:f_theta} also could be defined as 
\begin{equation}\label{eq:ODE_model}
    \hat{x}((k+1)T_\mathrm{s}) = \hat{x} (kT_\mathrm{s}) + I_m(f_\theta,T_\mathrm{s},\hat{x} (kT_\mathrm{s}),u (kT_\mathrm{s})).
\end{equation}


\textbf{Example 1.} Given the convergence and stability of the Runge-Kutta methods, they have been selected as the numerical integrator in this paper. For a $4^{th}$-order Runge–Kutta method we have: 
\begin{equation} \label{eq:4th_order_integration}
I_4 (f_\theta,h) =\sum_{i=1}^{4} w_{i,4} f_\theta(\hat{x}_{i,4},u_k) + \mathcal{O}(h^5),
\end{equation} where $h$ is the integration step, taken as $h = T_\mathrm{s}$.
The values of nodes, $\hat{x}_{i,4}$ are given by:
\begin{subequations} \label{eq:4_runge_kutta}
\begin{align} \hat{x}_{1,4} &= \hat{x}(kT_\mathrm{s}), \\
\hat{x}_{2,4} &= \hat{x}(kT_\mathrm{s}) + \frac{T_\mathrm{s}}{2} f_\theta(\hat{x}_{1,4}, u_k), \\
\hat{x}_{3,4} &= \hat{x}(kT_\mathrm{s}) + \frac{T_\mathrm{s}}{2} f_\theta(\hat{x}_{2,4}, u_k), \\
\hat{x}_{4,4} &= \hat{x}(kT_\mathrm{s}) + T_\mathrm{s} f_\theta(\hat{x}_{3,4}, u_k). 
\end{align} 
\end{subequations}
The corresponding weights, $w_{i,4}$, for this method are defined as:
\begin{equation}
    \begin{aligned}
        w_{1,4} &= \frac{T_\mathrm{s}}{6}, \
        w_{2,4} &= \frac{T_\mathrm{s}}{3}, \
        w_{3,4} &= \frac{T_\mathrm{s}}{3}, \
        w_{4,4} &= \frac{T_\mathrm{s}}{6}.
\end{aligned}
\end{equation}
Recalling \eqref{eq:one-step ahead}, we can rewrite  \eqref{eq:ODE_model} as
\begin{equation} \label{eq:modeled state}
    \hat{x}_{k+1} = {f}_{\theta,m}(\hat{x}_{k},u_{k}).
\end{equation}

Now that we have calculated the simulated states we can calculate the simulated outputs by recalling \eqref{eq:1_step ahead predictor}. Hence, for $n\geq 1$, we have:
\begin{equation}\label{eq:1step ahead predictor model}
    \hat{y}_k^{k+n} = {\Gamma_\theta}_{n,m} (\hat{x}_k, u_k^{k+n-1}).
\end{equation}

\section{Proposed method} \label{sec:proposed method}
Next, for the introduced model structure \eqref{eq: port_H_parametrized}, we introduce an efficient identification approach.
\subsection{Identification criterion} \label{sec:identification approach}
The classical approach to find the parameter vector $\theta$ in model set, given the collected input-outputs $\mathcal{D}_N$ is to minimize the $\ell_2$ norm of the prediction error $\hat{e}_k = y_k - \hat{y}_k$; which is achieved by minimizing the loss function defined as 
\vspace{-\abovedisplayskip}
\begin{equation}\label{eq:main_equations}
    V_{\mathcal{D}_N}(\theta)  = \frac{1}{N} \sum_{k=1}^{N} \left\| \hat{y}_k - y_k \right\|_2^2, 
\end{equation}
subjected to \eqref{eq: port_H_parametrized}.
As discussed in \ref{subsec:Model class}, here the OE model structure is chosen hence \eqref{eq:main_equations} expresses the simulation error loss function \cite{moradi2023physics}.

\subsection{Subspace encoder-based estimation}
To compute the loss function \eqref{eq:main_equations}, it is necessary to calculate the simulated outputs for all measured data, for which we can utilize \eqref{eq:1step ahead predictor model}. However, the computation of simulated outputs involves a significant number of operations in series, making it impractical for large datasets. Furthermore, the minimization process \eqref{eq:main_equations} can lead to numerous local minima and unstable optimization behavior when gradient-based techniques are used. To solve these problems, a subspace encoder method is proposed for the estimation of \eqref{eq: port_H_parametrized} through the minimization of \eqref{eq:main_equations} \cite{beintema2023continuous}. In this method, multiple truncated subsections are used to minimize the averaged prediction loss in a computationally efficient sense. In addition, an encoder estimates the initial state for each of these subsections which is crucial for the calculation of the prediction loss.

To estimate the initial state within a subsection containing $T$ samples, starting at $\tau\in[n+1,N-T]$ (denoted as $\hat{x}_{\tau|\tau}$), we introduce a parameterized encoder function by recalling  \eqref{eq:estimated reconstructability map}:
\begin{equation} \label{eq:parametrized encoder}
   \hat{x}_{\tau|\tau} = \psi_\eta (u_{\tau-n}^{\tau-1},y_{\tau-n}^{\tau}),
\end{equation}
where ${\psi_\eta}$ is an encoder function with encoder lengths $n$, parametrized with $\eta$. This function is formulated as an MLP in this paper. 
To distinguish between different subsections, we use the (current index \textbar{} start index) notation for clarity in indexing. Here, the \textit{current index} refers to the time step being evaluated within a given subsection. The \textit{start index} indicates the initial time step of that subsection from which the truncated data begins.

By implementing the subspace encoder \eqref{eq:parametrized encoder} and recalling \eqref{eq:modeled state} and \eqref{eq:1step ahead predictor model}, we can rewrite \eqref{eq:main_equations} as 
\begin{subequations}\label{eq:SUBNET_pHNN}
\begin{align}
    V^{\text{sub}}_{\mathcal{D}_N}(\theta,\eta) &= \frac{1}{C} \sum_{\tau=n+1}^{N-T+1}\sum_{k=0}^{T-1} \left\|y_{\tau+k}- \hat{y}_{\tau+k|\tau}\right\|_2^2, \label{subeq:loss_theta}\\
    \intertext{subject to:}
    \hat{x}_{\tau|\tau} &= \psi_\eta (u_{\tau-n}^{\tau-1},y_{\tau-n}^{\tau}), \label{subeq:x_theta}\\
    \hat{x}_{\tau+k+1|\tau} &= {f}_{\theta,m} (\hat{x}_{\tau+k|\tau}, u_{\tau+k}), \label{subeq:x_k+1_theta}\\
    \hat{y}_{\tau+k|\tau} &= h_{\theta} (\hat{x}_{\tau+k|\tau}), \label{subeq:y_theta}
\end{align}
\end{subequations}
 here, $C = (N - T - n + 1)T$. Fig. \ref{fig:port-HNN and SUBNET} illustrates the proposed approach in more details.  

The primary goal of minimizing the loss function \eqref{eq:SUBNET_pHNN} is to train the neural networks that form both the encoder function $\psi_\eta$ and are embedded within $f_{\theta,m}$ and $h_\theta$. These neural networks incorporate the port-Hamiltonian matrices: $H_\theta$, $J_\theta$, $R_\theta$, and $G_\theta$. As shown in Fig. \ref{fig:port-HNN and SUBNET}, the encoder neural network estimates the initial state, $\hat{x}_{\tau|\tau}$, for each subsection using past input and output data.  From this initial state, the state evolution is calculated iteratively using \eqref{subeq:x_k+1_theta}, and the corresponding outputs are computed via \eqref{subeq:y_theta}. This process is repeated across the entire subsection, generating simulated outputs that are then compared with measured outputs using the loss function \eqref{subeq:loss_theta}. This approach allows for the simultaneous training of all neural networks within the model. By limiting the simulation to shorter subsections, the computational cost is reduced, making the optimization process both efficient and scalable.

The process of minimizing the loss function \eqref{eq:SUBNET_pHNN} begins with the random initialization of all networks, as shown in Fig. \ref{fig:port-HNN and SUBNET}. After initialization, the loss is computed for the given normalized data (or batches of data), and the computation graph, including intermediate values, is stored in memory. The gradient of the loss is then computed via back-propagation using the stored computation graph. The network parameters are updated iteratively using a stochastic gradient optimization method, such as Adam \cite{kingma2014adam}. This procedure is repeated until the loss converges or early stopping is triggered based on cross-validation.

\section{Consistency analysis}
In this section, we analyze the consistency of the proposed method introduced in Section \ref{sec:proposed method}. Although the system under consideration \eqref{eq:system_class} is inherently continuous, we focus on the consistency of the proposed method when dealing with sampled outputs. We demonstrate this consistency for the discrete-time system and model, as detailed in Subsections \ref{subsec:Discretized-time system} and \ref{subsec:Discretized-time model}, respectively. Our consistency analysis builds upon the results from \cite{ljung1978convergence} and \cite{beintema2023deep}, with some results from \cite{beintema2023deep} recapitulated for completeness. Additionally, in Subsection \ref{subsec:Continuous-time consistency} we show that as the order of the numerical integrator tends to infinity, the consistency holds to the continuous-time system representation \eqref{eq:system_class} using Theorem \ref{thm:Convergence of numerical integration}.

\subsection{Data generating system}
We assume the measured data, $\mathcal{D}_N$, originates from a continuous-time dynamical system governed by \eqref{eq:system_class}. As discussed in Subsection \ref{subsec:Discretized-time system}, the system outputs can be reformulated using the 1-step ahead predictor introduced in \eqref{eq:1_step ahead predictor}. For the OE model structure, this predictor is equivalent to simulating the process dynamics starting from a given initial condition.  

For $k\geq 1$, let's define $\mathcal{E}_{[1,k]}$ as a $\sigma$-algebra generated by random variables $(u_1^k,v_1^k)$ with $\mu_\mathcal{E}:\mathcal{E}_{[1,k]}\rightarrow [0,1]$ as the probability mapping. 
Given the deterministic assumption over $f$ and $h$, and recalling \eqref{eq:compact}, we can introduce the set of all sampled trajectories of the data generating system \eqref{eq:compact} as 
\begin{equation}
    \begin{aligned}
             B = & \{ (y^{\infty}_1, x^{\infty}_1, u^{\infty}_1, v^{\infty}_1) \in (\mathbb{R}^{n_w})^N |\ (u^{\infty}_1, v^{\infty}_1) \in\mathcal{E}_{[1,\infty]}, \\
            & \text{ and } (y_k, x_k, u_k, v_k) \text{ satisfies } \eqref{eq:compact}  \ \forall k \in \mathbb{N} \},
    \end{aligned}
\end{equation}
where $n_w = n_y+n_x+n_u+n_y$. We represent the stochastic behavior of \eqref{eq:compact} by defining the $\sigma$-algebra $\mathcal{B}$ over $B$ and proper probabity $\mu_B$. Also, for a realization of the sample path for the time interval $[k_0,k]\subseteq \mathbb{N}$, we define $\mathcal{B}_{[k_0,k]}$. Given a sample path $\{(y_k, x_k, u_k, e_k)\}_{k=k_0}^{\infty} \in B_{[k_{0}, \infty]}$ of \eqref{eq:compact} with the initial state $x(k_0) = x_0$, the sequence $\{(\tilde{y}_k, \tilde{x}_k, u_k, e_k)\}_{k=k_0}^{\infty} \in B_{[k_{0}, \infty]}$ represents the response of \eqref{eq:compact} when the initial state is perturbed to $\tilde{x}(k_0) = \tilde{x}_0$ at time $k_0 \in \mathbb{N}$, while being subjected to the same input and disturbance as the original state response.

\textbf{Condition 1}\label{c1} (\textit{Output exponential stability}) The data-generating system \eqref{eq:compact} is incrementally exponentially output stable, if, $\forall \delta>0$, there exist a $0\leq C(\delta)<\infty$, and $0<\lambda\leq 1$, such that
\begin{equation}\label{eq:stability}
   \mathbb{E} [\| y_{k} - \breve{y}_{k} \|^{4}_{2}] < C(\delta)\lambda^{k-k_0}, \quad \forall k \geq k_0
\end{equation}
for any $k_0\geq 1$, $x_0, \breve{x}_0 \in \mathbb{R}^{n_x}$ with $\|x_0 - \breve{x}_0\|_2^2 < \delta $, and $(u_1^\infty, v_1^\infty)\in \mathcal{E}_{[1,\infty]}$ where the random variables $y_k$ and $\breve{y}_k$ belong to $\mathcal{B}_{[k,\infty]}$ with the same $(u_k,v_k)$, but $x_{k_0}=x_0$, and $\breve{x}_{k_0}=\breve{x}_0$.

\subsection{Model set}\label{subsec:Model set}
The model introduced in \eqref{subeq:x_theta}-\eqref{subeq:y_theta} corresponds to a model structure $M_\xi$ with finite-dimensional parameter vector $\xi =[\theta^\top \quad \eta^\top] ^\top$, constrained within a compact set $\Xi \subset \mathbb{R}^{n_{\xi}}$. Here, the model set is $M = \{M_{\xi} \mid \xi \in \Xi\}$; for each $ \xi \in \Xi$, the OE-pHNN model $M_\xi$ with a given encoder lag $n\geq 1$, can be written as a one-step ahead predictor:
\begin{equation}\label{eq:parametrized predictor}
    \hat{y}_{\tau+k|\tau} = \hat{\gamma}_{k,m}(\xi; {y}^{\tau+k-1}_{\tau-n}, u^{\tau+k-1}_{\tau-n}).
\end{equation}
For $M$, two key conditions must be satisfied. First, the 1-step ahead predictor introduced in \eqref{eq:parametrized predictor} must be differentiable with respect to $\xi$, as specified in Condition 2. Second, to ensure the convergence of the predictor, $\hat{\gamma}_{k,m}$, Condition 3 requires that the effects of delayed inputs and outputs diminish exponentially.

\textbf{Condition 2} \label{C2}(\textit{Differentiability}) The 1-step ahead predictor $\hat{\gamma}_{k,m}: \mathbb R^{n_\theta+n_\eta} \times \mathbb R^{(n_y+n_u)(n+k)} \rightarrow \mathbb R^{n_y}$ in \eqref{eq:parametrized predictor} is diffrentiable w.r.t. $\xi$ on $\Bar{\Xi}$ which is an open neighborhood of $\Xi$.

\textbf{Condition 3} \label{C3} (\textit{Predictor convergence}) For a differentiable predictor $\hat{\gamma}_{k,m}$ as described in Condition 2, there exist a $0\leq C<\infty$ and $0\leq \lambda <1$, such that $\forall k\geq 0$ and $\xi \in \Bar{\Xi}$:
\begin{equation} \label{eq: C3}
\begin{aligned}
    & \left\| \hat{\gamma}_{k,m}(\xi; u^{k-1}_{-n}, y^{k-1}_{-n}) - \hat{\gamma}_{k,m}(\xi; \breve{u}^{k-1}_{-n}, \breve{y}^{k-1}_{-n}) \right\|_2^2 \\
    &  \leq C \sum_{s=-n}^{k} \lambda^{k-s} (\left\| u_s - \breve{u}_s \right\|_2^2 + \left\| y_s - \breve{y}_s \right\|_2^2).
\end{aligned}
\end{equation}
for any $(u^{k-1}_{-n}, y^{k-1}_{-n}), (\breve{u}^{k-1}_{-n}, \breve{y}^{k-1}_{-n}) \in \mathbb R^{(n_y+n_u)(n+k)}$ and $\| \hat{\gamma}_{k,m}(\xi, 0^{k-1}_{-n}, 0^{k-1}_{-n}) \|_2^2 \leq C$, where $0_{\tau-n}^{\tau+k-1} = [0...0]^{\top}$. 

The same condition is necessary for
$\frac{d}{d\xi} \hat{\gamma}_{k,m}(\xi; y^{k-1}_{-n}, u^{k-1}_{-n})$.

\begin{thm} \label{thm:Convergence} (\textit{Convergence}) For the data-generating system \eqref{eq:compact}, under Condition 1 and quasi-stationary input $u$, 
if the set of models $M$ defined by \eqref{subeq:x_theta}-\eqref{subeq:y_theta} for all $\xi$ in $\Xi$ fulfill Condition 2 and 3, then 
\begin{equation}\label{eq: convergence}
    \sup_{\xi \in \Xi} \left\| V_{\mathcal{D}_N}^{\text{sub}}(\xi) - \mathbb{E}[V^{\text{sub}}_{\mathcal{D}_N}(\xi)] \right\|_2^2 \rightarrow 0
\end{equation}
with probability 1 as $T, N \rightarrow\infty $.
\end{thm} 
\textbf{Proof. }For the proof see \cite{ljung1978convergence}. $\quad \blacksquare$

Via \eqref{eq: convergence}, we ensure the uniform convergence of $V_{\mathcal{D}_N}^{\text{sub}}(\xi)$ to its expected value, $\mathbb{E}[V^{\text{sub}}_{\mathcal{D}_N}(\xi)]$ as the dataset size increases indefinitely. This convergence is essential for consistent parameter estimation, as it links finite-sample performance to the asymptotic behavior of the system.

\subsection{Consistency}\label{subsec:consistency}
According to the definition in \cite{ljung1978convergence},  consistency means that the model becomes asymptotically unbiased and converges to an accurate representation of the system. This property ensures that with sufficient data, the model will reliably capture the behaviour of the system.

To show consistency, it is necessary to assume that the system is part of the model set. Hence, momentary assume that \eqref{eq:one-step ahead} and \eqref{eq:recurrent} generated the data $\mathcal{D}_N$.
Consider the state-reconstructability map $\psi_n$ in \eqref{subeq:psi} for \eqref{eq:compact} with $n \geq n_x$. Note that 
\begin{align}
    y_{\tau+k} &= \gamma_{k,m}(\psi_n(u^{\tau-1}_{\tau-n}, y^{\tau}_{\tau-n}-v_{\tau-n}^\tau), u^{\tau+k-1}_{\tau}) + v_{\tau+k}, \\
            &= \breve{\gamma}_{k,m}(u^{\tau+k-1}_{\tau-n}, y^{\tau+k-1}_{\tau-n}-v_{\tau-n}^\tau) + v_{\tau+k} 
\end{align}
 $\forall \tau,k \geq 0$, where $\gamma_{k,m}$ is defined in \eqref{eq:1_step ahead predictor}, i.e., $\gamma_{k,m} = (h \circ_k {f_{d,m}})$. Hence, we have
 \begin{equation}
     \bar{y}_{\tau+k|\tau} = \mathbb{E}[\breve{\gamma}_{k,m}(u^{\tau+k-1}_{\tau-n}, y^{\tau+k-1}_{\tau-n}-v_{\tau-n}^\tau)] = \bar{\gamma}_{k,m}(u^{\tau+k-1}_{\tau-n}, y^{\tau+k-1}_{\tau-n}),  
 \end{equation}
which is the optimal one-step-ahead predictor associated with \eqref{eq:compact} under an $n$-lag based reconstructability map. To be able to reconstruct the state in the general case, we need to assume $n$ is large enough, i.e., $n \geq 2n_x+1$; ${f}_{\theta,m}$ is Lipschitz continuous in $x$, and the output function, $h_\theta$, has a finite amount of nondegenerate critical points, $x_i$, with $h_\theta(x_i) \neq h_\theta(x_i)$ for $i\neq j$ \cite{murray2013existence,jose_internship}.

\textbf{Definition 1} (\textit{Equivalence set}) To ensure the considered model sets are sufficiently rich to encompass an equivalent model of the data-generating system \eqref{eq:compact}, we define a set of models as
\begin{equation}\label{eq: equivalent set}
 \Xi^* = \{\xi \in \Xi \,|\, \hat{\gamma}_{k,m}(\xi, \mathbf{.}) \equiv \bar{\gamma}_{k,m}(\mathbf{.}), \forall k \geq 0\}   \end{equation}
such that $ \Xi^*\neq \varnothing$. Accordingly, we can ensure the existence of an  $M_{\xi^*} \in M$ that corresponds to data-generating system \eqref{eq:compact}.

\textbf{Condition 4} (\textit{Persistency of excitation}) We define input sequence $u_1^N$ in the collected data $\mathcal{D}_N$ as weekly persistent when $\forall \xi_1 , \xi_2 \in \Xi$ for which $V^{sub}_{(.)}(\xi_1) \neq V^{sub}_{(.)}(\xi_2)$, the inequality
\begin{equation}
    V^{sub}_{\mathcal{D}_N}(\xi_1) \neq V^{sub}_{\mathcal{D}_N}(\xi_2)
\end{equation}
holds with a probability of 1.

\textbf{Property 1} (\textit{Minimal cost}) To show consistency, it is essential that every element in $\Xi^*$ possesses minimal cost corresponds to \eqref{eq:SUBNET_pHNN}. Hence we need to show that $\forall \xi\in \Xi$ and $ \xi^* \in \Xi^*$ for the loss function \eqref{eq:SUBNET_pHNN}, we have:
\begin{equation}
    \lim_{\substack{T,N\rightarrow\infty }} V^{sub}_{\mathcal{D}_N} (\xi^*) \leq \lim_{\substack{T,N\rightarrow\infty }} V^{sub}_{\mathcal{D}_N} (\xi),
\end{equation}
with probability 1.

\textbf{Proof. } Given the existence of $ \mathbb{E}[V^{sub}_{\mathcal{D}_N}(\xi^*)] $ by Theorem \ref{thm:Convergence}, we need to show that $\forall \tau\in[1,N] $:
\begin{equation}
    \lim_{{T \to \infty}} \frac{1}{T} \sum_{{k=0}}^{T-1} \left\| y_{{\tau+k}} -\hat{y}^*_{{\tau+k|\tau}} \right\|_2^2 \leq \lim_{{T \to \infty}} \frac{1}{T} \sum_{{k=0}}^{T-1} \left\| y_{{\tau+k}} - \hat{y}_{{\tau+k|\tau}} \right\|_2^2
\end{equation}
with $\hat{y}^*_{{\tau+k}|\tau} = \hat{\gamma}_{k,m}(\xi^*; y^{\tau+k-1}_{\tau-n}, u^{\tau+k-1}_{\tau-n})$, which is proven in \cite{beintema2023continuous}. $\quad \blacksquare$

\begin{thm}\label{thm:Consistency}(\textit{Consistency}) Under the Theorem \ref{thm:Convergence}, Condition 4 and Property 1:
\begin{equation}
    \lim_{\substack{T,N\rightarrow\infty }} \tilde{\xi}_N \in \Xi^*
\end{equation}
with probability 1, where
\begin{equation}
    \tilde{\xi}_N = \text{arg}\min_{\xi\in\Xi} {V^{sub}_{\mathcal{D}_N}(\xi)}.
\end{equation}
\end{thm} 
\textbf{Proof.} See \cite{ljung1978convergence}. $\quad \blacksquare$

\subsection{Continuous-time consistency}\label{subsec:Continuous-time consistency}
Having established consistency for the discretized system with a numerical integrator of finite order $m$, i.e., Equation \eqref{eq:compact}, we now consider consistency as the integrator order tends to infinity, i.e., $m\rightarrow\infty$.

\begin{thm} \label{thm:Convergence discrete-time system} \textit{(Convergence of the discrete-time system output to continuous-time output)} Theorem \ref{thm:Convergence of numerical integration} enables us to reformulate the output of the system \eqref{eq:compact} as follows
\begin{equation} \label{eq:infinity_1step ahead predictor model}
\lim_{m \rightarrow \infty} {y}_k^{k+n} = {\Gamma}_{n} ({x}_k, u_k^{k+n-1}).
\end{equation}
where ${\Gamma}_{n}$ represents the $n$-step continuous-time output mapping of the system. 
\end{thm}

\textbf{Proof.} From Theorem \ref{thm:Convergence of numerical integration}, as $m$ increases to infinity, the relative error between the exact integral, $I(f,T_\mathrm{s},x_k,u_k)$ in \eqref{eq:integral_system} and the approximation $I_m(f,T_\mathrm{s},x_k,u_k)$ in \eqref{eq:ODE_system} vanishes. 
Therefore, we can write the following limit for the discrete-time state transition \eqref{eq:one-step ahead} 
\begin{equation}
    \lim_{m \rightarrow \infty} x_{k+1} = f_d(x_k,u_k).
\end{equation}

By introducing this result into the recursive output equations  \eqref{eq:recurrent}, and subsequently \eqref{eq:compact}, we establish that as $\smash{m\rightarrow\infty}$, the discretized output mapping $\Gamma_{n,m}$ converges to the continuous-time output mapping $\Gamma_{n}$. This guarantees that the output of the discrete-time system, $y_{k}^{k+n}$, is equivalent to the continuous-time output trajectory sampled at $t = kT_\mathrm{s}$.$\quad \blacksquare$

\textit{Extension to Continuous-Time Consistency:}
Based on the Theorem \ref{thm:Convergence discrete-time system}, the response of the discrete-time system \eqref{eq:compact} converges to the output of the continuous-time system \eqref{eq:infinity_1step ahead predictor model} for the sampled outputs at $t=kT_\mathrm{s}$ as $m\rightarrow\infty$.  Given the consistency of the discrete-time system proven in Theorem \ref{thm:Consistency}, and the convergence established by the Theorem \ref{thm:Convergence discrete-time system}, the consistency result extends to the continuous-time system. This shows that as the integrator order grows, the model remains consistent with the continuous-time system dynamics \eqref{eq:infinity_1step ahead predictor model}, ensuring reliable identification of the underlying physical system.

\textbf{Remark.} Given the assumption of Gaussian white noise for the measurement noise, as $T \rightarrow \infty$, the quadratic loss function in \eqref{eq:SUBNET_pHNN} results in a maximum likelihood estimator that is asymptotically efficient \cite{ljung1995system}.

\section{Simulation results}
This section demonstrates the performance of the proposed OE-pHNN identification approach through detailed simulation studies. 
\subsection{Data-generating system} \label{subsec:data_generation}
\begin{figure}[t]
    \centering
    \includegraphics[scale=0.4]{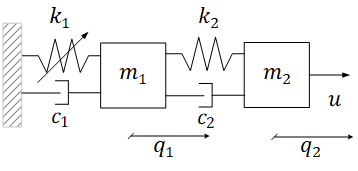}
    \caption{Schematic representation of two body-based connected mass-spring-damper systems.}
    \label{fig:oscillator}
\end{figure}

We simulate two coupled mass-spring-damper (MSD) systems, as illustrated in Fig. \ref{fig:oscillator};
In this system, $m_i$, $k_i$, and $c_i$ represent the mass, spring stiffness and damper coefficient of the $i^{th}$ oscillator, respectively. $q_i(t)$ denotes the position of the corresponding mass, and $u(t)$ is the applied force to the system. Here, it is assumed the first spring is cubic i.e. $k_1(q) = k_0+k_{c}q^2$. The parameters of the system are set as follows: $m_1=m_2=1$  kg, $k_0=k_2=1$N/m, $k_c =0.1$ N/m\(^3\), and $c_1=c_2=0.5$ Ns/m. 

\subsection{Dataset and model training} \label{subsec:dataset and model}
For gathering experimental data, a multisine excitation is used since it provides full control over the applied power spectrum \cite{pintelon2012system}. The considered input is defined as \vspace{1mm}
\begin{equation}\label{u_input}
    u(t) = \sum_{k=1}^{100} \sin{(2\pi kf_0t+\phi_{k})},
\end{equation}
where $f_0 = 0.01$ and the phase components $\phi_k$ are randomly chosen within the interval $[0, 2\pi)$. The output of the system is the sampled velocity of the second mass, $\dot{q}_2$, with a sampling time, $T_\mathrm{s} = 0.1$  seconds.  For the training and validation datasets, the measured output is contaminated by Gaussian white noise $e_k$, with a signal-to-noise ratio (SNR) from 50dB to 35dB.

To generate the datasets for our experiments, we created 48 different input realizations and recorded the response of the system for each, starting from random initial states uniformly distributed over the state space. The simulations were performed using the explicit Runge-Kutta method of order 8, and the data were collected at different sampling rates:
\begin{enumerate}
    \item The first dataset $(u_k, y_k)$, sampled at the sampling rate $T_\mathrm{s}$ was used for training, validation, and testing. This dataset was split into training, validation, and test sets using a fixed ratio of 5:2:5, respectively.
    \item The second dataset, $(\bar{u}_k, \bar{y}_k)$, sampled at a finer interval $T_{\text{fine}} = \frac{T_\mathrm{s}}{n}$ with $n \geq 1$, represented the ground truth behavior of the system at a higher resolution. This dataset  was reserved for later testing to evaluate the performance of the model against higher-resolution data.
\end{enumerate} 

\begin{figure}
    \centering
    \includegraphics[width=0.85\linewidth]{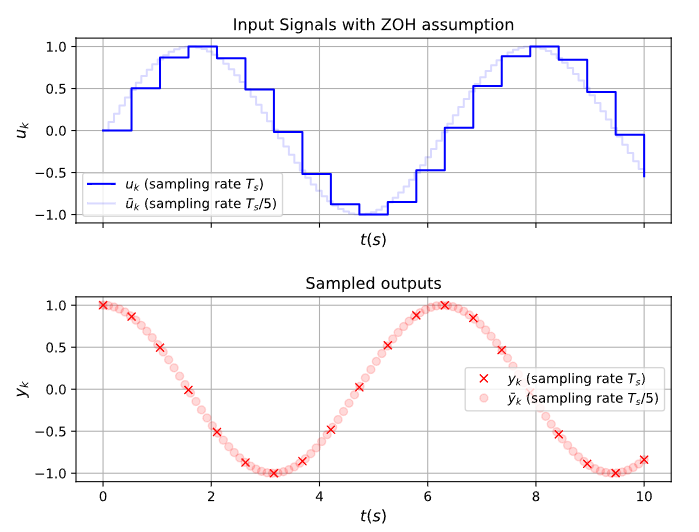}
    \caption{Sampled inputs and outputs with sampling time $T_\mathrm{s}$, $(u_k,y_k)$ vs sampling time $\frac{T_\mathrm{s}}{5}$, $(\bar{u}_k,\bar{y}_k)$.}
    \label{fig:sampled input and output}
\end{figure}

Fig. \ref{fig:sampled input and output} provides a schematic overview of the input and output data in these datasets.

For model training, we used hyperparameters chosen based on preliminary experimentation to achieve a balance between model complexity and convergence rate. Specifically, the model order was set as $n_x=4$, and $n_a = n_b = 20$ \cite{beintema2023continuous}. The truncation length, $T$, varied for each model, as detailed in Fig. \ref{fig:NRMS_N}.

The functions $J_\theta$, $R_\theta$, and $G_\theta$ were implemented using a multi-layer perceptron (MLP) with a single hidden layer containing 8 nodes. The Hamiltonian function, $H_\theta$, was modeled as a feedforward neural network with two hidden layers, each containing 16 nodes. For the encoder networks, a default configuration of two hidden layers with 64 nodes each was employed. The activation function used for all networks was the hyperbolic tangent ($\tanh$). Training was performed using the Adam optimizer \cite{kingma2014adam}, with a batch size of 256 and a learning rate of $10^{-3}$.

The accuracy of the simulation is characterized via \emph{Normalized Root Mean Square error} (NRMS)
\begin{equation} \label{eq:NRMS}
\text{NRMS} = \frac{\text{RMS}}{\sigma_y} = \frac{\sqrt{\frac{1}{N} \sum_{k=1}^{N} \| y_k - \hat{y}_k \|_2^2}}{\sigma_y}.
\end{equation}
where $\hat{y}_k$ represents the simulated output of the OE-pHNN model, and $y_k$ is the measured output in the considered data set. Here, $\sigma_y$ is the sample standard deviation of $y$.

\subsection{Performance of the OE-pHNN}
\begin{figure}[t]
    \centering
    \includegraphics[width=0.9\linewidth]{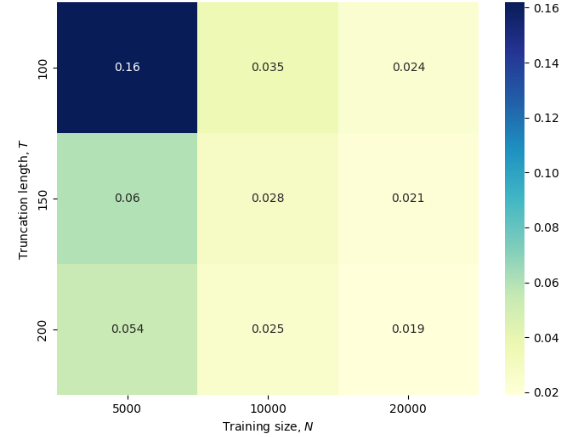}
    \caption{NRMS of the simulated model response on the test dataset for the MSD system, when the model is estimated under different training sample sizes, $N$, and truncation lengths $T$, while SNR=50dB.}
    \label{fig:NRMS_N}
\end{figure}

To evaluate the performance of the proposed OE-pHNN approach, we compared the NRMS of the simulated model response on the test dataset across different combinations of training dataset sizes $N$, and truncation lengths $T$ under SNR=50dB. 
Fig. \ref{fig:NRMS_N} \footnote{The implementation of the simulation study is available at \url{https://github.com/sarvin90/OE-pHNN}} presents a heatmap of NRMS values across different training sizes and truncation lengths, providing insights into their impact on model accuracy. The results indicate that increasing both $N$ and $T$ generally improves model accuracy. This aligns with the theoretical consistency guarantees described in Theorem \ref{thm:Consistency}. For smaller training sizes (e.g., $N=5,000$) and shorter truncation lengths (e.g., $T=100$), the NRMS is higher, reflecting reduced accuracy due to insufficient data. In contrast, larger datasets with extended truncation lengths (e.g., $N=20,000$ and $T=200$) achieve minimum NRMS values, showing the capacity of the model to generalize and accurately simulate behaviour of the system.

To assess the effect of measurement noise, the accuracy of the simulated results was evaluated on a dataset with a training size of 20,000 and a truncation length of 200. As the noise level increased, the accuracy slightly decreased but remained within an acceptable range. These results are detailed in Table \ref{table:NRMS_Noise_Specific}.


\begin{table}
    \centering
    \caption{NRMS of the simulated model response on the test dataset for the MSD system under different measurements noise levels for $N=20,000$ and $T=200$.} \label{table:NRMS_Noise_Specific}
    \begin{tabular}{|c|c|} 
        \hline
        SNR (dB) & NRMS \\ \hline
        50              & 0.019 \\ 
        40              & 0.023 \\ 
        35              & 0.025 \\ \hline
    \end{tabular}
\end{table}

\subsection{Higher resolution test dataset}
To validate the performance of the trained model, we evaluated its ability to simulate system behavior at higher sampling rates using a refined dataset $(\bar{u}_k, \bar{y}_k)$ as described in Subsection \ref{subsec:data_generation}. This enables to assess whether the trained model has captured the underlying continuous-time dynamics of the system. During this evaluation process, the model is trained with sampling time $T_\mathrm{s} = 0.1$ seconds, but tested at a faster sampling rate (higher resolution) $T_{\mathrm{fine}} = \frac{T_\mathrm{s}}{n}$, $n\geq 1$.

\begin{figure}
    \centering
    \includegraphics[width=\linewidth]{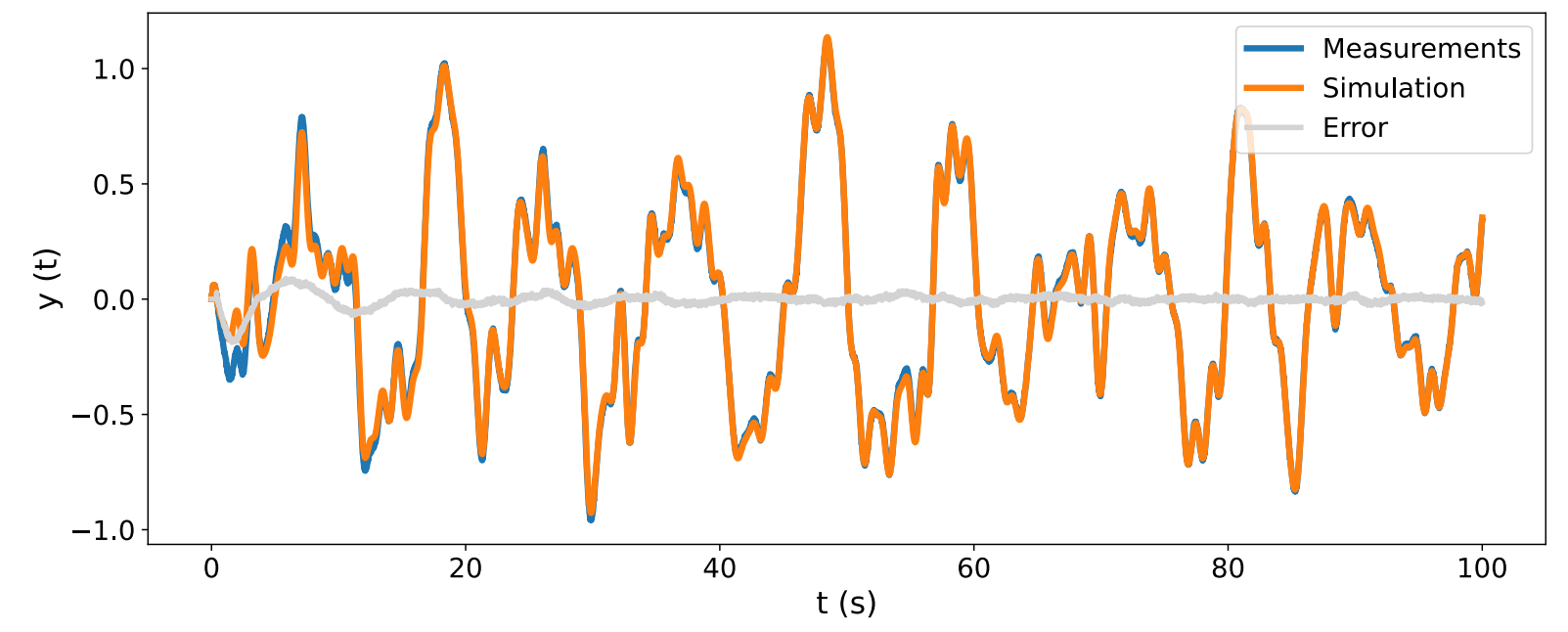}
    \caption{Measurements vs simulation for $\frac{T_\mathrm{s}}{T_{fine}}=5$  (first 100 seconds) for the MSD system.}
    \label{fig:ratio_5}
\end{figure}

The accuracy of the simulated response of the model for the higher resolution dataset with a 5 times faster sampling rate is evident in Fig. \ref{fig:ratio_5}. This shows that we have successfully identified the underlying continuous-time system dynamics, allowing us to use our model at different (higher) sampling resolutions without losing accuracy. However, a closer examination of the error reveals that the performance of the model is less accurate at the beginning. This discrepancy is due to the encoder function, which is only valid at the sampling frequency $T_\mathrm{s}$ for which it was trained. While just for initialization, resampling the data to match the trained sampling frequency would enable the encoder to be used at other sampling rates, this falls outside the scope of this work.

\begin{figure}
    \centering
    \begin{subfigure}[b]{0.4\textwidth}
        \centering
        \includegraphics[width=\linewidth]{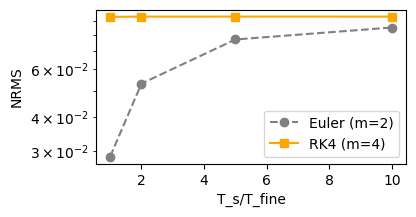} 
        \caption{The model trained by Euler numeric integrator.}
        \label{fig:Euler}
    \end{subfigure}
    \begin{subfigure}[b]{0.38\textwidth}
        \centering
        \includegraphics[width=\linewidth]{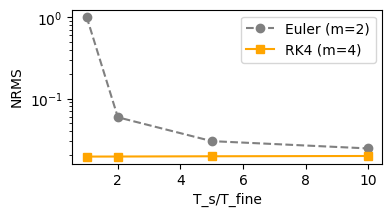} %
        \caption{The model trained by RK4 numeric integrator.}
        \label{fig:RK4}
    \end{subfigure}
    \hfill
    \caption{Comparison of model performance across different test dataset resolutions and integrators.}
    \label{fig:combined_plots}
\end{figure}

To further evaluate the impact of the numerical integrator on the model performance, we tested the Euler method within the OE-pHNN approach during both training and testing. As shown in Table \ref{table:NRMS_combined}, the model trained with the Euler method provides acceptable results only when tested with the Euler method at the same data resolution used during training; the error increases when considering test data with faster sampling rates, see also Fig. \ref{fig:Euler}. Increasing the order of the numerical integrator during testing (switching from Euler to RK4) does not improve the results. The results suggest that the model exhibits an overfitting behavior, compensating for the low accuracy of the numerical integrator during training. This overfitting explains why switching to a higher-order integrator, such as RK4, fails to enhance performance. This indicates that the model failed to capture the true continuous-time behavior of the system during training.

In contrast, a model trained with the RK4 numerical integration method obtains high-quality results for all sampling rates in the test datasets. It can also be observed that the Euler-based test error of the RK4 trained model converges towards the RK4 test error for higher sampling rates as shown in Fig. \ref{fig:RK4}. Indeed, one can expect that the approximation error of the Euler integration becomes smaller for smaller sampling times.
This illustrates, as indicated in the consistency discussion in Subsection~\ref{subsec:Continuous-time consistency}, that the numerical integration approach should be of sufficient high order to accurately identify the underlying continuous-time system dynamics.

\begin{table}
    \centering
    \caption{NRMS of the simulated model response on the test dataset for the MSD system when tested under various sampling rates, $T_{fine}$, and numerical formulas. The model is trained using either the Euler or RK4 numerical integration method.}
    \label{table:NRMS_combined}
    \begin{tabular}{|c|c|c|c|c|}
        \hline
        \multirow{2}{*}{Train numeric integrator} & \multirow{2}{*}{$\frac{T_\mathrm{s}}{T_{fine}}$} & \multicolumn{2}{c|}{Test numeric integrator} \\ \cline{3-4}
                   &  & Euler  & RK4  \\ \hline
        \multirow{4}{*}{Euler ($m=2$)}  
                   & 1  & 0.02860  & 0.09340 \\  
                   & 2  & 0.05308  & 0.09358 \\  
                   & 5  & 0.07711  & 0.09361 \\  
                   & 10 & 0.08534  & 0.09360 \\ \hline        
        \multirow{4}{*}{RK4 ($m=4$)}  
                   & 1  & 1.02574  & 0.01926 \\  
                   & 2  & 0.05878  & 0.01929 \\  
                   & 5  & 0.02986  & 0.01943 \\  
                   & 10 & 0.02427  & 0.01959 \\ \hline        
    \end{tabular}
\end{table}

\section{Benchmark result}
In addition to the simulation example, the proposed method has been tested on an identification benchmark with real experimental data, showing its practical applicability and robustness. This section discusses the cascaded tanks benchmark system \cite{schoukens2016cascaded}, a nonlinear setup exhibiting weak nonlinearity under normal conditions, with a hard saturation effect during high input signal peaks.
\begin{figure} [h]
    \centering 
    \includegraphics[scale=0.6]{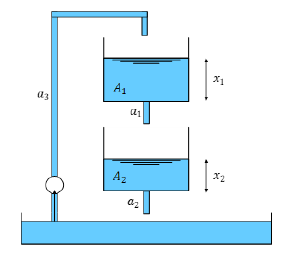}  
    \caption{Schematic of the two cascaded tanks.}
    \label{fig:cascaded_schematic}
\end{figure} 

\subsection{Data description}
This benchmark is a cascaded tanks system, where a pump controls the water flow from a reservoir into an upper tank, which then flows into a lower tank and back into the reservoir as depicted in Fig. \ref{fig:cascaded_schematic}. The nonlinear characteristics of the system are observed in the water flow between the tanks. here the input is the pump voltage, and the output is the water
level of the lower tank $x_2$. The benchmark includes two datasets, each containing 1024 samples with sampling rate of $T_\mathrm{s} = 4$s. The first dataset is used for training, while the first 512 samples of the second dataset are reserved for validation. The entire second dataset is utilized for testing.

\subsection{Applying the OE-pHNN identification method}
In this system, we treat the height of each tank as a state, thus $n_x = 2$ is chosen. Here, $n_a = n_b = 4$, and $T=60$; these hyperparameters are selected based on the analysis presented in  \cite{beintema2023continuous}. For all neural networks, we use an MLP with one hidden layers and 8 nodes with tanh activation function. For training, the Adam optimizer is used with a learning rate of $10^{-3}$ and a batch size of 64.

\subsection{Result}
In this section, we compare the performance of the OE-pHNN approach with other identification methods. When comparing the results in Table \ref{tab:identification_methods_rmse}, it should be noted that while the total data available remains unchanged, different methods may have used varying ratios for splitting the training, validation, and test sets. As shown in Table \ref{tab:identification_methods_rmse}, our model achieves a test RMS of 0.28, placing it among the more effective methods for identifying this system. This illustrates that despite the limited data record available, our model demonstrates robust performance. Additionally, we can guarantee the stability of our model due to the inherent dissipativity of port-Hamiltonian systems.

\begin{figure}
    \centering
    \includegraphics[width=\linewidth]{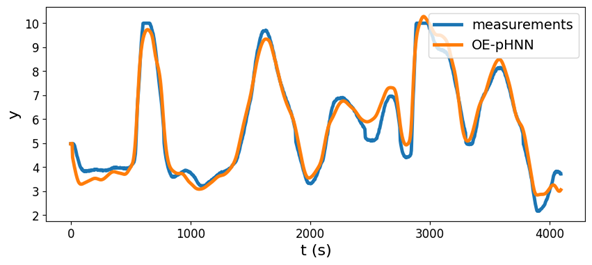}
    \caption{Simulation vs output measurements of the height of the second tank.}
    \label{fig:sim_cascade}
\end{figure}

Fig. \ref{fig:sim_cascade} illustrates the simulation results, effectively capturing the nonlinearity in the sytem. This visualization confirms that our model can accurately represent complex system behaviors, which is crucial for reliable identification.

\begin{table}
    \centering
    \caption{RMS of the simulated responses of estimated models for the cascaded-tank benchmark by various methods on the test data set. }
    \label{tab:identification_methods_rmse}
    \begin{tabular}{|l|c|}
        \hline
        \textbf{Identification Method} & \textbf{Test RMS} \\ \hline
        LTI SS \cite{champneys2024baseline}& 0.59\\
        RNN \cite{champneys2024baseline} & 0.54\\
        LSTM \cite{champneys2024baseline} & 0.49 \\
        State-space with GP prior \cite{svensson2017flexible} & 0.45 \\ 
        SCI \cite{forgione2021continuous} & 0.40 \\ 
        IO stable CT ANN \cite{weigand2023input} & 0.39 \\ 
        DT subspace encoder \cite{beintema2023continuous} & 0.37 \\ 
        TSEM \cite{forgione2021continuous}  & 0.33 \\ 
        Tensor B-splines \cite{karagoz2020nonlinear} & 0.30 \\ 
        \hline
        \textbf{OE-pHNN (this study)} & \textbf{0.28} \\ 
        \hline
        CT subspace encoder \cite{beintema2023continuous}& 0.22 \\  
        Grey-Box \cite{rogers2017grey} & 0.19 \\
        \hline
    \end{tabular}
    
\end{table}

\section{Conclusion}
In this paper, we have proposed a novel approach that enhances Hamiltonian neural networks (HNNs) by integrating port-Hamiltonian theory and output-error (OE) modeling. This integration incorporates physics knowledge by embedding the energy exchange and dissipation properties of physical systems into the neural network structure. Our OE-pHNN framework effectively manages external inputs, dissipation, and measurement noise, utilizing deep subspace encoding through the SUBNET method for efficient dynamic model identification. This method has proven to be both consistent and effective, as evidenced by its superior performance in simulation examples and benchmark results. By addressing noisy measurements and enhancing model accuracy, our work provides a reliable and scalable solution for complex nonlinear system identification.  

\bibliographystyle{plain}        
\bibliography{autosam}           

\end{document}